\documentclass[11pt]{article}

\usepackage{authblk}
\usepackage{times}
\usepackage{soul}
\usepackage{url}
\usepackage[hidelinks]{hyperref}
\usepackage[utf8]{inputenc}
\usepackage[small]{caption}
\usepackage{graphicx}
\usepackage{amsmath}
\usepackage{amsthm}
\usepackage{multirow}
\usepackage{mathtools}
\usepackage{enumerate}
\usepackage{booktabs}
\usepackage{varwidth}
\usepackage{latexsym}
\usepackage{amssymb}
\usepackage{booktabs}
\usepackage{amsmath}
\usepackage{todonotes}
\DeclareMathOperator*{\argmax}{argmax}

\begin{document}

\date{}

\title{Inferring spatial relations from textual descriptions of images}

\author[1]{Aitzol Elu}
\author[1]{Gorka Azkune}
\author[1]{Oier Lopez de Lacalle}
\author[2, 3, 4]{Ignacio Arganda-Carreras}
\author[1]{Aitor Soroa}
\author[1]{Eneko Agirre}

\affil[1]{
HiTZ Basque Center for Language Technologies - Ixa NLP Group, University of the Basque Country UPV/EHU, M. Lardizabal 1, Donostia 20018, Basque Country, Spain}
%\email{aelu003@ikasle.ehu.eus}

\affil[2]{
Dept. of Computer Science and Artificial Intelligence, University of the Basque Country (UPV/EHU), Paseo Manuel Lardizabal 1, 20018 Donostia-San Sebastian, Spain}

\affil[3]{
Ikerbasque, Basque Foundation for Science, Maria Diaz de Haro 3, 48013 Bilbao, Spain}

\affil[4]{
Donostia International Physics Center (DIPC), Paseo Manuel Lardizabal 4, 20018 Donostia-San Sebastian, Spain}

\affil[ ]{\textit {aelu003@ikasle.ehu.eus,\{gorka.azcune,oier.lopezdelacalle,ignacio.arganda,a.soroa, e.agirre\}@ehu.eus}}
%\email{gorka.azcune@ehu.eus}

\maketitle

\begin{abstract}
Generating an image from its textual description requires both a certain level of language understanding and common sense knowledge about the spatial relations of the physical entities being described. In this work, we focus on inferring the spatial relation between entities, a key step in the process of composing scenes based on text. More specifically, given a caption containing a mention to a subject and the location and size of the bounding box of that subject, our goal is to predict the location and size of an object mentioned in the caption. Previous work did not use the caption text information, but a manually provided relation holding between the subject and the object. In fact, the used evaluation datasets contain manually annotated ontological triplets but no captions, making the exercise unrealistic: a manual step was required; and systems did not leverage the richer information in captions. 
Here we present a system that uses the full caption, and \emph{Relations in Captions} (REC-COCO), a dataset derived from MS-COCO which allows to evaluate spatial relation inference from captions directly. Our experiments show that: (1) it is possible to infer the size and location of an object with respect to a given subject directly from the caption; (2) the use of full text allows to place the object better than using a manually annotated relation. Our work paves the way for systems that, given a caption, decide which entities need to be depicted and their respective location and sizes, in order to then generate the final image.
\end{abstract}

%\keywords{Text-to-image synthesis;  Natural Language Understanding; Spatial relations; Deep learning}




%%


%%%%%%%%%%%%%%%%%%%%%%%%%%%%%%%%%%%%%%%%%%%%%%%%%%%%%%%%
%% INTRODUCTION
%%%%%%%%%%%%%%%%%%%%%%%%%%%%%%%%%%%%%%%%%%%%%%%%%%%%%%%%

\section{Introduction}
\label{sec:introduction}
The ability of automatically generating images from textual descriptions is a fundamental skill which can boost many relevant applications, such as art generation and computer-aided design. From a scientific point of view, it also drives research progress in multimodal learning and inference across vision and language, which is currently a very active research area \cite{mogadala2019trends}. In the case of scenes comprising several entities, it is necessary to infer which is an adequate scene layout, i.e., which entities to show, their location and size. 

From the language understanding perspective, in order to generate realistic images from textual descriptions, it is necessary to infer visual features and relations between the entities mentioned in the text. For example, given the text \textit{"a black cat on a table"}, an automatic system has to understand that the cat has a certain color (\textit{black}) and is situated on top of the table, among other details. In this paper, we focus on the spatial relations between the entities, since they are the key to suitably compose scenes described in texts. The spatial information is sometimes given explicitly, in form of prepositions (\textit{"cat on a table"}), but more often implicitly, since the verb used to relate two entities contains information about the spatial arrangement of both. For example, from the text (\textit{"a woman riding a horse"}) it is obvious for humans that the woman is on top of the horse. However, acquiring such spatial relations from text is far from trivial, as this kind of common sense spatial knowledge is rarely stated explicitly in natural language text~\cite{vandume2010}. That is precisely what text-to-image systems learn, relating both explicit and implicit spatial relations expressed in text with actual visual arrangements showed in images. 

\begin{figure}[t]
\centerline{\includegraphics[height=3in, width=3.5in]{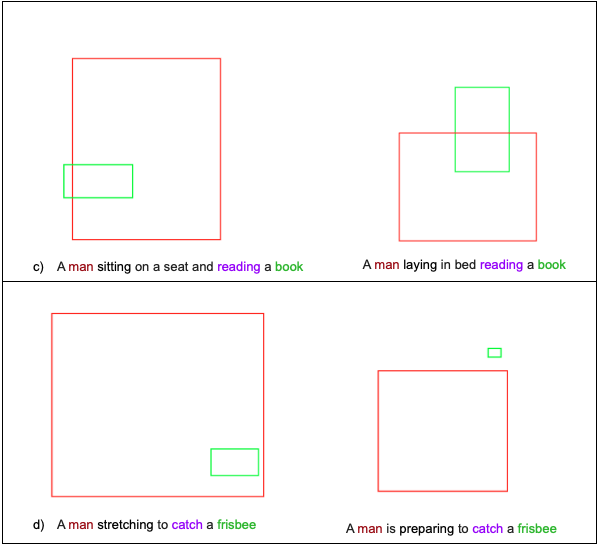}}
\caption{An example to illustrate the relevance of full captions in spatial relation inference. Given a caption, a subject token in the caption, the bounding box for the subject (both in red), and a target object in the caption (in green), the systems need to return the bounding box for the object (see Figure 2 for the actual images). The relationship between subject and object is highlighted in purple. In each row we can see two different layouts for the same subject, object and relation, motivating the need to model the full caption. Best viewed in color. See Figure \ref{errorfig} for the actual images.}\label{problem}
\end{figure}

\begin{figure}[t]
\centerline{\includegraphics[height=3in, width=3.5in]{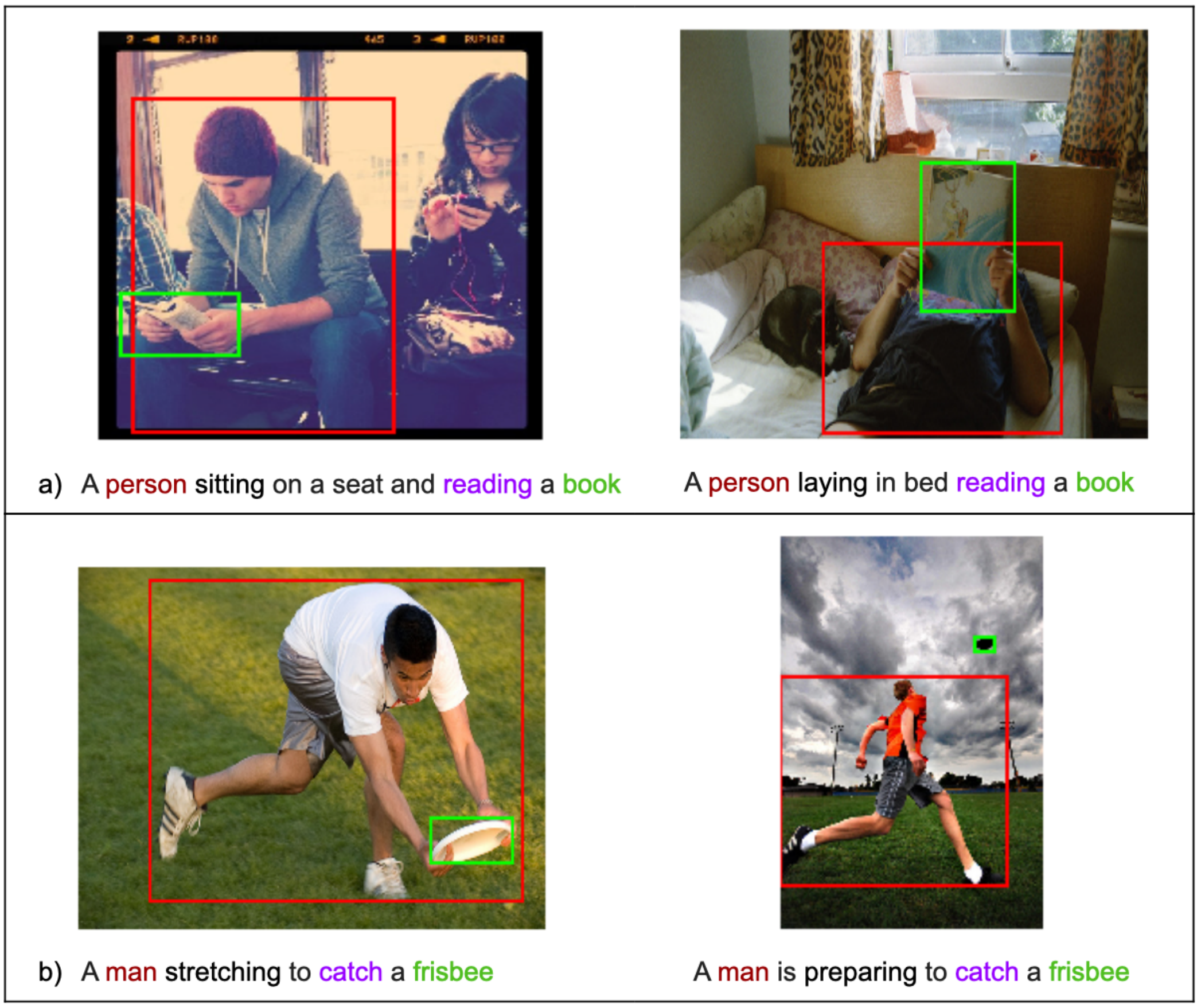}}
\caption{The images that underlie the bounding boxes in Figure 1. Best viewed in color.} \label{errorfig}
\end{figure}

A large strand of research in text-to-image generation are evaluated according to the pixel-based quality of the generated images and the global fidelity to the textual descriptions, but do not evaluate whether the entities have been arranged according to the spatial relations mentioned in the text \cite{reed2016generative}. Closer to our goal, some researchers do focus on learning spatial relations between entities ~\cite{bagherinezhad2016elephants,collell2018acquiring,gupta2015visual,huang2019sketchforme,Krishna2017,malinowski2014pooling}. For instance, in \cite{gupta2015visual,Krishna2017} the authors proposed to associate actions along with their semantic arguments (subject and object) with pixels in images (i.e., bounding boxes of entities) as a way towards understanding the images. V-COCO is a dataset which comprises images and manually created Subject, Relation, Object ($S,R,O$) ontological triplets, henceforth called \emph{concept triplets}, where each $S$ and $O$ is associated with a bounding box in the image \cite{gupta2015visual}. Note that the terms used to describe the triplet concepts are selected manually from among a small vocabulary\footnote{In this paper we will use uppercase words for ontology concepts, as opposed to lowercase for caption words.} of an ontology, e.g. \emph{PERSON} or \emph{BOOK}, and are not linked to the words in the caption. Visual Genome is constructed similarly \cite{Krishna2017}. 
Typically, those datasets are created by showing images to human annotators, and asking them to locate the bounding boxes of the entities participating on predefined relations, and to select the terms for the relation and entities from a reduced vocabulary in a small ontology. Using such a dataset, \cite{collell2018acquiring} presents a system that uses concept triplets to infer the spatial relation between the subject $S$ and the object $O$. Given the bounding box of the subject, the system outputs the location and size of the bounding box of the object. Evaluation is done checking whether the predicted bounding box matches the actual bounding box in the image. The datasets and systems in the previous work require the use of manually extracted ontological triplets, and systems did not use the actual captions, posing two issues: a manual pre-processing step was required; and systems did not use the richer information in captions.

In this paper we propose to \textbf{study the use of full captions instead of manually selected relations when inferring the spatial relations between two entities}, where one of them is considered the \textit{subject} and the other is the \textit{object} of the action being described by the \textit{relation}. The problem we address is depicted in Figure \ref{problem}. Given a textual description of an image and the location (bounding box) of the subject of the action in the description, we want the system to predict the bounding box of the target object. Note that we do not use the actual pixels for this task, but we include Figure \ref{errorfig} for illustrative purposes. To the best of our knowledge, there is no previous work addressing the same problem, i.e. nobody studied before whether using full captions instead of concept triplets benefits spatial relation inference. 

Our \textbf{hypothesis} is that the textual description\footnote{In this paper we use caption and textual description interchangeably.} accompanying the image contains information that helps inferring the spatial relations of two entities. We argue that the information presented in manually created triplets alone is often insufficient to properly infer spatial relations. As a motivation, Figure \ref{errorfig} shows pairs of examples (left and right) where the relation between the subject and the object (given by a verb) is not enough to correctly predict the spatial relation between them. In each row there are two examples for the same subject, relation and object (e.g. \emph{person}, \emph{reading}, \emph{book}), but the spatial relation between subject and object is different, and depends on the interpretation of the rest of the caption. For instance, in the top-left caption the person is sitting while it is reading a book, so that the book is around the middle of the bounding box for the person, while in the top-right caption the person is laying in bed, and therefore the book is slightly above the person.

To validate the main hypothesis of our work, we created a new dataset called \emph{Relations in Captions} (REC-COCO) that contains associations between caption tokens and bounding boxes in images. REC-COCO is based on the MS-COCO~\cite{lin2014microsoft} and V-COCO datasets~\cite{gupta2015visual}. For each image in V-COCO, we collect their corresponding captions from MS-COCO and automatically align the concept triplet in V-COCO to the tokens in the caption. This requires finding the token for concepts such as \emph{PERSON}.  As a result, REC-COCO contains the captions and the tokens which correspond to each subject and object, as well as the bounding boxes for the subject and object (cf. Figure \ref{rec-coco-adibidea}). 

In addition, we have adapted a well-known state-of-the-art architecture that worked on concept triplets \cite{collell2018acquiring} to work also with full captions, and performed experiments which show that: 
(1) It is possible to infer the size and location of an object with respect to a given subject directly from the caption; (2) The use of the full text of the caption allows to place the object better than using the manually extracted relation. 

The main contributions of the work are the following:

\begin{itemize}
    \item We show for the first time that the textual description includes information that is complementary to the relation between a subject and an object. From another perspective, our work shows that, given a caption, a reference subject and an object in the caption, our system can assign a location and a size to the object using the information in the caption, without any manually added relation.
    \item We introduce a new dataset created for this task. The dataset comprises pairs of images and captions, including, for each pair, the tokens in the caption that describe the subject and object, and the bounding boxes of subject and object. The dataset is publicly available under a free license\footnote{https://github.com/ixa-ehu/rec-coco}.
\end{itemize}

%%%%%%%%%%%%%%%%%%%%%%%%%%%%%%%%%%%%%%%%%%%%%%%%%%%%%%%%
%% RELATED WORK
%%%%%%%%%%%%%%%%%%%%%%%%%%%%%%%%%%%%%%%%%%%%%%%%%%%%%%%%

\section{Related work}

Understanding the spatial relations between entities and their distribution in space is essential to solve several tasks such as human-machine collaboration \cite{guadarrama2013grounding} or text-to-scene synthesis ~\cite{hinz2019generating,huang2019sketchforme,jyothi2019layoutvae}, and has attracted the attention of different research communities. In this section, we will provide the different approaches to infer spatial relations among entities, evaluation methodologies arisen from those communities and available resources such as datasets.

\textbf{Visual scene understanding.} 
There has been a great interest in tasks related to visual scene understanding in recent years, such as human-object interaction, semantic segmentation or object detection. As a consequence, there are large-scale image-based datasets like MS-COCO \cite{lin2014microsoft}, V-COCO \cite{gupta2015visual} or Visual Genome \cite{Krishna2017}.% or HICO-DET \cite{chao:wacv2018}. 
Those datasets contain very rich and diverse scenes combining humans and their daily environments, accompanied by textual descriptions and/or structured text, among others. Thus, in principle, they should be appropriate to test whether textual descriptions are useful to infer spatial relations between entities.

However, none of those datasets combine concept triplets, image descriptions, textual triplets as mentioned in the textual description, and the bounding boxes of the subject and object for each instance. V-COCO is the most similar, but the captions and the mentions of the concepts as expressed in the caption are not included. We thus had to build a new dataset, REC-COCO, which contains all that information.

\textbf{Spatial common sense knowledge.} Initial proposals created rule-based systems to generate spatial representations \cite{Kruijff2007}. With the arrival of deep learning systems this task began to gain more interest among researchers. Malowinski et al.  \cite{malinowski2014pooling} demonstrated that it was possible to create a system to estimate spatial templates from structured input such as (Object$_1$, spatial$\_$preposition, Object$_2$)  \cite{platonov2018computational}. Collell et al. \cite{collell2018acquiring} proposed the task of predicting the 2D relative spatial arrangement of two entities under a relationship given a concept triplet (Subject, Relation, Object). The template is determined by the interaction/composition of the Subject, Relation and Object, so changing one of the concepts that make up the structured input may change the spatial template. Contrary to those previous works,  we  argue  that  the  information  presented  in the concept triplets alone is often insufficient to properly infer spatial relations. Therefore, we propose to check whether textual descriptions in the form of captions encode contextual information which is useful to infer spatial relations and thus place entities better in an image. %We show that the use of the caption in addition to the textual triplet improves over the manual concept triplets, and the improvement also holds when the relation is not explicitly given to the system.

\textbf{Text-to-image synthesis.} Recent studies have proposed a variety of models to generate an image given a sentence. Reed et al. \cite{reed2016generative} used a GAN \cite{goodfellow2014generative} that is conditioned on a text encoding for generating images of flowers and birds. Zhang et al. \cite{Xu_2018_CVPR} proposed a GAN based image generation framework where the image is progressively generated in two stages at increasing resolutions.  Reed et al. \cite{NIPS2016_6111} performed image generation with sentence input along with additional information in the form of keypoints or bounding boxes. Some works \cite{48087,li2019objectdriven} break down the process of generating an image from a sentence into multiple stages.  The input sentence is first used to predict the entities that are presenting the scene,  followed  by the prediction of bounding boxes, then semantic segmentation masks, and finally the image. These works are aligned with ours, since they also assume that the spatial relations can be obtained from paired textual descriptions and images, as we do. However, their focus is on image generation and they do not prove that using raw textual information is actually helpful for spatial relation inference. In that sense, our work provides a solid foundation for their design choices and thus, complements their work.

\textbf{Quantitative information about entities.} There is a line of work to determine the quantitative relation between two nouns on a specific scale \cite{forbes-choi-2017-verb,yang-etal-2018-extracting}. These types of relations are key for image understanding tasks such as image captioning \cite{silberer_uijlings_lapata_2018,wang2020learning} and visual question answering cite{aditya2019spatial,bai2020decomvqanet}. The common theme in the recent work \cite{davidov2010extraction,narisawa2013204,tandon2014acquiring} is to use search query templates with other textual cues (e.g., more than, at least, as many as, and so on), collect numerical values, and model sizes as a normal distribution. However, the quality and scale of such extraction is somewhat limited. Bagherinezhad et al.  \cite{bagherinezhad2016elephants} showed that textual observations about the relative sizes of entities are very limited, and relative size comparisons are better collected through visual data. In this sense, our work shows that it is possible to extract information about the relative sizes of entities, learning the implicit relations that appear in the raw text.
In \cite{DBLP:journals/corr/abs-1906-01327} the authors automatically collected large amounts of web data and created a resource with distributions over physical quantities that can be used to acquire common knowledge such as relative sizes of entities, but they did not use images for that goal. Their work is complementary to ours, as we use multimodal data instead of textual alone. 

\textbf{Other multimodal tasks.} Following the success of transformers in natural language processing \cite{devlin2018bert}, multimodal transformers have been proposed to tackle several multimodal tasks with similar architecture designs. Good examples are ViLBERT \cite{lu2019vilbert}, VisualBERT \cite{li2019visualbert} and InterBERT \cite{lin2020interbert}. Those multimodal transformers have shown strong performance in multimodal tasks such as visual question answering, visual commonsense reasoning, natural language for visual reasoning and region-to-phrase grounding. However, multimodal transformers have been investigated for discriminative tasks, rather than generative tasks such as image generation. Only very recently a solution has been proposed for text-to-image generation: X-LXMERT \cite{cho2020xlxmert}, which shows that multimodal transformers can also generate state-of-the-art images from textual input. For that purpose, authors proposed to sample visual features for masked inputs and to add an image generator to transform those sampled visual features into images. Although the proposal is very relevant for the field, the suggested solution does not explicitly model the spatial layout of entities and thus, it cannot be used for the purposes of this work. 

%%%%%%%%%%%%%%%%%%%%%%%%%%%%%%%%%%%%%%%%%%%%%%%%%%%%%%%%
%% REC-COCO DATASET
%%%%%%%%%%%%%%%%%%%%%%%%%%%%%%%%%%%%%%%%%%%%%%%%%%%%%%%%

\section{REC-COCO Dataset}
\label{sec:recCOCO}

The main goal of this paper is to extract spatial relations among the entities mentioned in image captions. To the best of our knowledge, there exists no dataset that contains explicit correspondences between image pixels (bounding boxes of entities) and their respective mentions in the image descriptions. We thus developed a new dataset, called \emph{Relations in Captions} (REC-COCO), that contains such correspondences. REC-COCO is derived from MS-COCO~ \cite{lin2014microsoft} and V-COCO~\cite{gupta2015visual}. The former is a collection of images, each image described by 5 different captions. The latter comprises a subset of the MS-COCO images, where each image has a manually created Subject, Relation, Object ($S, R, O$) concept triplet, a bounding box for the subject and a bounding box for the object. V-COCO uses ontology concepts to describe the elements of ($S,R,O$) triplets, e.g. (PERSON, READ, BOOK). The triplets correspond to actions performed by the subject on the object. Given the bounding box of the subject and the triplet, the dataset has been used to evaluate whether the system has been able to infer the spatial relation between the subject and object and thus produce the correct bounding box for the object. Note that the concept triplets are not linked to the actual words used in the image captions. 

\begin{figure}[t]
\centerline{\includegraphics[width=\textwidth]{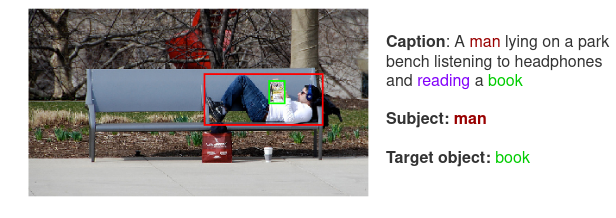}}
\caption{Example of REC-COCO. Given the caption, the subject token (\emph{man}), the bounding box for the subject (in red), and the target object (\emph{book}), the systems need to return the bounding box for the object (in green). The dataset is automatically created from V-COCO and MS-COCO, matching the ontological triplet in V-COCO (\textcolor{red}{PERSON},\textcolor{violet}{READ},\textcolor{green}{BOOK}) with the tokens in the MS-COCO captions. The actual image is included for illustration purposes, it is not used by the systems. Best viewed in color.} 
\label{rec-coco-adibidea}
\end{figure}

In order to be able to access the information in the captions, we devised an automatic method to map MS-COCO and V-COCO, so that each term in the V-COCO ($S,R,O$) triplet is linked to the most similar token on one of the MS-COCO captions for the corresponding image. If the similarities between terms and tokens is below a threshold the example is discarded. To create the links, the method considers all ($S,R,O$) triplets of V-COCO images in turn. For each triplet and image, it first gathers all five captions from MS-COCO, and represents each caption by concatenating the normalized vector embeddings of each word. Let $\mathbf{C^{i}} = [\mathbf{c}_{1}^{i};\ldots;\mathbf{c}_{N_{i}}^{i}]$ be the matrix representing caption $i$ ($i \in [1,5]$), where column $\mathbf{c}_{j}^{i}$ is the unit normalized embedding of the $j$-th word. Let also ($\mathbf{s}, \mathbf{r}, \mathbf{o}$) be the unit normalized embeddings of the terms used to describe the elements in the ($S,R,O$) triplet from V-COCO. For each caption $i$, the algorithm first obtains the word in the caption  ($j^{i}$) that is closest to each of the embeddings of the concept triplet, as well as the similarity score between them ($\mathrm{sc}^i$). For example, the method would compute $j^{i}_{s}$ and $\mathrm{sc}_S^i$ for the $S$ element in the triplet, as follows:
\begin{eqnarray*}
  \mathrm{sc}_S^i &=& \max \mathbf{s}^T\cdot\mathbf{C}^i  \\
  j^{i}_{S} &=& \argmax_{j \in |\mathbf{C}^i|}\mathbf{s}^T\cdot\mathbf{C}^i
\end{eqnarray*}

\noindent $j_R^i, \mathrm{sc}_R^i, j_O^i, \mathrm{sc}_O^i$ are calculated likewise for $R$ and $O$. Afterwards, it selects the caption $i$ whose sum of scores is maximum:

\begin{eqnarray*}
  i &=& \argmax \mathrm{sc}_S^i+\mathrm{sc}_R^i+ \mathrm{sc}_O^i
\end{eqnarray*}

If the similarity score is below certain threshold\footnote{The threshold was empirically set to $0.75$ based on manual inspection of the resulting dataset.}, the triplet is discarded. If not, the caption word corresponding to index $j_S^i$  (respectively $j_R^i, j_O^i$) is selected to represent the subject of the triplet (same for relation and object). 

By applying the method described above, each element of the ($S,R,O$) triplets in V-COCO is anchored to actual words occurring in captions accompanying the image. Figure \ref{rec-coco-adibidea} shows a sample output, where the subject and object in the V-COCO triplet (PERSON, READ, BOOK) are linked to the corresponding words in the MS-COCO caption, \emph{man} and \emph{book}, respectively. In addition, the action concept READ is matched to the token \emph{reading}. We discarded V-COCO triplets corresponding to actions that do not explicitly require a subject and an object, and that have a single argument instead (\textit{smile}, \emph{look}, \emph{stand}, and so on). All in all,  REC-COCO comprises $19,559$ instances from $6,407$ different images. Each instance consists of an image, a caption, the subject and object words and the bounding boxes of subject and object. In addition, in order to enable contrastive experiments, the V-COCO concept triplet and the word corresponding to the relation are also provided in the dataset. 
Table \ref{statistics_table} shows further statistics of the dataset.

\begin{table}
\centering
\begin{tabular}{lr}
%\toprule\\[-7pt]
\hline
Number of Images & $6,407$\\
Number of Captions & $14,928$\\
Number of Instances & $19,559$\\
Captions per Image    & $2.33$\\ % \pm1.26$\\ %-              & 5       & -            & 1 \\
Subject-Object Pairs per Caption &  1.31 \\ 
\hline
%\\[-6pt]\toprule
\end{tabular}
\caption{Statistics of the REC-COCO dataset. Note that an image can be described by more than one caption, and that a caption can contain more that one subject-object pair of interest. } %*means and standard deviation.}
\label{statistics_table}
\end{table}

As the method to generate the dataset is automatic, we checked the quality of the produced alignments by manually annotating 100 random samples of the dataset. For each subject and object pair extracted by the automatic alignment algorithm, we checked whether the tokens matched the action described by the concept triplet. The results can be seen in Table \ref{tab-matching-precision}. More than 85 examples got either the subject or the object correctly aligned with caption tokens, and in 71 examples both of them were correctly aligned. In addition, we also checked whether the verb describing the action could be correctly identified. Identifying the token that describes the relation is more difficult, and the algorithm is only able to do it correctly for 56 examples. 

\begin{table*}[htbp]%\scriptsize
 \begin{center}
  \begin{tabular}{lc}
   \hline
    Term & Accuracy\\
   \hline
   Subject & 86\% \\
   Object & 85\% \\
   Subject \& Object & 71\% \\
%   Relation & 56\% \\ % % kasu NIL daude.
   \hline
  \end{tabular}
  \caption{Quality of token identification in REC-COCO for 100 random samples.}
  \label{tab-matching-precision}
 \end{center}
\end{table*}

%%%%%%%%%%%%%%%%%%%%%%%%%%%%%%%%%%%%%%%%%%%%%%%%%%%%%%%%
%% MODELS FOR INFERRING SPATIAL RELATIONS
%%%%%%%%%%%%%%%%%%%%%%%%%%%%%%%%%%%%%%%%%%%%%%%%%%%%%%%%

\section{Model for inferring spatial relations from captions}
\label{sec:models-inferr-spat}

The problem addressed in this work  (cf. Figure \ref{rec-coco-adibidea}) is the following: given a caption, a subject token in the caption ($S$), the location and size of the bounding box for the subject, and a target object ($O$), the system needs to predict sensible location and size of the bounding box for the object. 
More formally, we denote as $O^c = [O_x^c, O_y^c] \in \mathbb{R}^2$ the $(x,y)$ coordinates of the center of the bounding box covering the object $O$, and  $O^b = [O_x^b, O_y^b] \in \mathbb{R}^2$ half of its width and height. Thus, we use $O = [O^c, O^b] \in \mathbb{R}^4$ as the ground truth location and size of the object. Model predictions are denoted with a hat $\widehat{O^c} , \widehat{O^b}$. The task is then to produce the location and size $\widehat{O} = [\widehat{O^c} , \widehat{O^b}] \in \mathbb{R}^4$ of the token filling the \emph{Object} role in the caption describing the scene given the bounding box of the \emph{Subject}, which is defined analogously $S = [S^c, S^b] \in \mathbb{R}^4$. 

\begin{figure*}[t!]
\centerline{\includegraphics[width=\columnwidth]{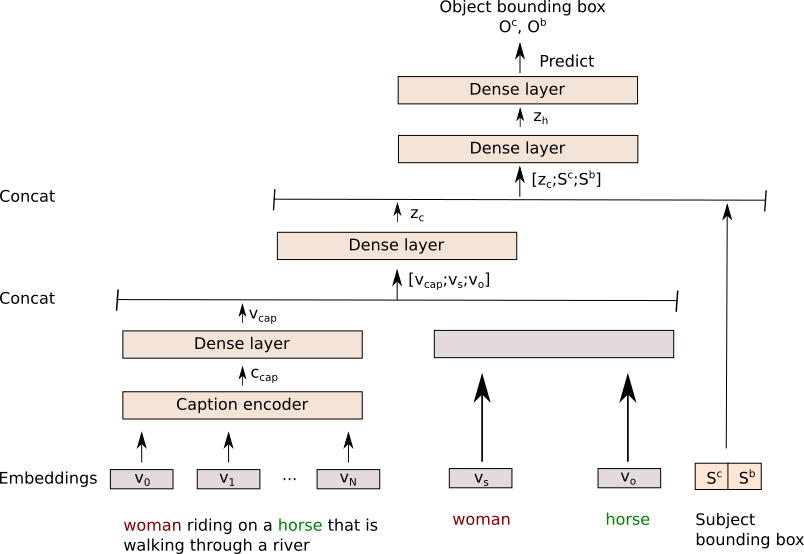}}
\caption{Architecture of the spatial relation inference model. The system receives as input a caption, the subject and object tokens, the location and size of the subject bounding box, and outputs the location and size of the subject. See text for further details.
} \label{modelfig}
\end{figure*}

The proposed model is a neural network inspired in \cite{collell2018acquiring}. We chose this model because of the excellent results in spatial relation inference, and adapted it to include the caption text in the input. Our model takes as input  the embeddings of the caption words, additional embeddings for subject and object tokens ($S$ and $O$), denoted respectively as $v_S$ and $v_O$, and the bounding box of the subject $[S^c, S^b]$. Figure \ref{modelfig} shows the diagram of the model, with input in the lower part and output in the top. The system first uses a caption encoder and a dense layer to produce the fixed-length representation of the caption. We tried different alternative caption encoders (see below). The output of the dense layer is concatenated to the embeddings of subject and object, and fed into a dense layer which encodes the caption, the subject and object tokens. This representation is concatenated to the subject bounding box representation and fed into the final dense layer, which is used to predict the object bounding box. 

We experimented with three different caption encoders in our experiments:

\paragraph{Average embedding (AVG)} This encoder just averages the embeddings of each token in the caption:

\begin{eqnarray*}
c_{\mathrm{cap}} &=& \frac{1}{N} \sum_{i=1}^{N} v_i
\end{eqnarray*}

where $v_i$ is the embedding of the $i$th word in the caption of length $N$.

\paragraph{BiLSTM encoder} The caption words are fed into a bidirectional LSTM \cite{graves2013speech} and the final hidden states of the left and right LSTMs are concatenated:

\begin{eqnarray*}
c_{\mathrm{cap}} &=& [h_N^{L} ; h_N^{R}]
\end{eqnarray*}

The embedding layer of the LSTM modules are initialized with external word embeddings, and the rest of weights are learned during training.

\paragraph{BERT encoder} In this setting we use a pre-trained BERT model~\cite{devlin2018bert}. More specifically, the caption of length $N$ is represented by the embedding corresponding to the special \textit{[CLS]}  token (position 0). BERT weights are fine-tuned during training.

\begin{eqnarray*}
c_{\mathrm{cap}} &=& BERT[0](v_0, \ldots, v_N)
\end{eqnarray*}

Given the output of any of the above caption encoders $c_{\mathrm{cap}}$, we stack a dense layer to obtain the final caption representation $v_{\mathrm{cap}}$:

\begin{eqnarray*}
v_{\mathrm{cap}} &=& \mathrm{ReLU}(W_{\mathrm{cap}}c_{\mathrm{cap}} + b_{\mathrm{cap}})\\
\end{eqnarray*}

The caption representation $v_{\mathrm{cap}}$ is then concatenated to the object and subject embeddings and fed into a dense layer to obtain the final textual embedding: 

\begin{eqnarray*}
z_c &=& ReLU(W_c[v_{\mathrm{cap}};v_S;v_O] + b_c)\\
\end{eqnarray*}

This representation is concatenated to the subject bounding box and a final regression dense layer produces the object bounding box:

\begin{eqnarray*}
z_h &=& ReLU(W_h[z_c;S^c; S^b] + b_h)\\
\widehat{O} &=& W_{\mathrm{out}}z_{h} + b_{\mathrm{out}}
\end{eqnarray*}

where $W_{\mathrm{cap}}, b_{\mathrm{cap}}, W_c, b_c, W_h, b_h, W_{\mathrm{out}}$ and $b_{\mathrm{out}}$ are the parameters of the model (along with the parameters of the caption encoders). We used the $ReLU$ activation function because it is widely used in similar neural network architectures. The loss function is the mean squared error  loss between the predicted and the actual values:
\begin{eqnarray*}
L(O,\widehat{O}) &=& \lVert \widehat{O} - O \rVert ^2
\end{eqnarray*}

%%%%%%%%%%%%%%%%%%%%%%%%%%%%%%%%%%%%%%%%%%%%%%%%%%%%%%%%
%% EXPERIMENTS
%%%%%%%%%%%%%%%%%%%%%%%%%%%%%%%%%%%%%%%%%%%%%%%%%%%%%%%%

\section{Experiments}
\label{sec:experiments}

In this section we report the results of the performed experiments. We conduct several sets of experiments, depending on the research question addressed. In the first set we assess the validity and quality of the REC-COCO dataset, complementing the analysis presented in Section \ref{sec:recCOCO}. In the second set, we study which encoder is the most effective for solving this task. In a third set, we check whether it is possible to infer the size and location of an object with respect to a given subject directly from the caption without the need of manually extracted concept triplets. In addition, we present a fourth set to study how complementary is the information in the captions with respect to the triplets. 

\begin{table*}[t]

\begin{center}
\begin{tabular}{|l|r|}
\hline
Caption Encoder & minutes per epoch \\
\hline
  AVG & 0.016 \\
  BiLSTM & 1.783 \\
  BERT & 9.783 \\
\hline
\end{tabular}
\caption{Average training time per epoch for our model when training on the REC-COCO dataset, depending on the caption encoder used. }
\label{tab-train-time}
\end{center}
\end{table*}

The evaluation metrics used within the paper are the ones proposed by Collel et al.~\cite{collell2018acquiring}, and include the following: {Above/below Classification Accuracy}, a binary metric that measures whether the model correctly predicts that the object center is above/below the subject in the image, where we  report both  macro averaged accuracy (acc$_y$) and macro-averaged F1 score ($F1_y$); {Pearson Correlation (r)} of both ($x,y$) axes between the predicted value and the ground truth; {Coefficient of Determination} ($R^2$) of the prediction and the ground truth \cite{draper1998applied}; {Intersection over Union (IoU)}, a bounding box overlap measure  \cite{everingham2015pascal}.

The data is preprocessed using the same procedure presented in~\cite{collell2018acquiring}, namely, we normalize the bounding box coordinates with the width and height of the images and apply a mirror transformation on the vertical axis to the image when the object is at the left of the subject. Textual captions are lowercased and punctuation marks are removed. 

Regarding model hyperparameters, we used $300$ dimension GloVe embeddings~\cite{Pennington14glove:global} that are publicly available\footnote{\url{http://nlp.stanford.edu/projects/glove}} to initialize all word embeddings used in the model. Regarding training details we use $10$-fold cross-validation to train all the models using $10$ epochs, a batch size of $64$ and a learning rate of $0.0001$ with an RMSprop optimizer. The same parameters are used when training the BiLSTM sentence encoders. We use default parameters when fine-tuning the BERT encoder and we trained for $5$ epochs with a batch size of $2$. All the experiments have been performed in a single Nvidia Titan XP GPU. The observed training time per epoch on the REC-COCO dataset is depicted in Table \ref{tab-train-time}. Most of the complexity lays in the training of the BiLSTM and BERT caption encoders. In fact, the rest of the model only takes 0.016 minutes per epoch, as shown by the training time needed by the average caption encoder.

\subsection{Assessing REC-COCO as a dataset}

In this set of experiments we want to assess the quality and validity of the REC-COCO dataset. More concretely, we want to check two important features of REC-COCO:
\begin{enumerate}
    \item The effect of the token alignment algorithm in the task: we check whether the noise introduced by the  token-concept alignment method used to create REC-COCO has any negative effect when comparing the results of a system running on the aligned tokens with respect to the results of a system running on the manual concept triplets. %(cf. Section \ref{sec:recCOCO}. As the automatic algorithm introduces noise, we want to assess the impact of that noise in the task.
    \item The difficulty of the task: the proposed task should be feasible to be resolved by automatic methods, yielding results which should be comparable to related datasets. %but at the same time it should not be too easy, in the sense that baseline methods can obtain good results.
\end{enumerate}

\begin{table}[t]
  \centering
{
\begin{tabular}{|c|c|cccccc|}
\hline
Dataset                                                     & Size   & $acc_y$       & $F1_y$      & $r_x$         & $r_y$         & $R^2$           & IoU\\
\hline

REC-COCO & 19,559 & {77.9} & {77.7} & 70.4         & {67.6}  & 47.3          & 12.1 \\
  \hline

V-COCO (subset) & 19,559 & 75.6          & 75.4         & {78.3} & 63.4      &  {51.7} & {14.9} \\
\hline
\multirow{2}{*}{Visual Genome}                              & 20,000 & 71.7          & 71.2         & 87.2          & 76.5          & 46.9           & 6.8 \\
\cline{2-8} 
                                                            & 378k   & 74.5          & 74.5         & 89.2          & 83.2          & 64.8           & 11.1 \\
\hline
\end{tabular}
}
\caption{Assessing REC-COCO as a dataset. Comparison of results attained on comparable datasets. V-COCO contains manually created ontological triplets and REC-COCO uses their automatically linked mentions. The results for V-COCO refer to the subset which was linked in REC-COCO. Visual Genome also contains images and manually annotated triplets, where annotations do not overlap with those in V-COCO or REC-COCO. All results use the same simplified model, where the relation was used instead of the full caption. }
\label{dataset_table}
\end{table}

In these experiments we train and run a model which is a simplified version of our full model, as it does not use the caption in the input, just the subject, relation and object. The architecture is the same as in Section \ref{sec:models-inferr-spat} (cf. Figure \ref{modelfig}), where instead of the output of the linear layer over the caption encoder $v_{\mathrm{cap}}$ we use the embedding of the relation $v_R$: 

\begin{eqnarray*}
z_c &=& ReLU(W_c[v_R;v_S;v_O] + b_c)\\
\end{eqnarray*}

We thus compare the results of this system under the same conditions across three different datasets: 

\paragraph{V-COCO} This dataset contains ($S,R,O$) concept triplets, e.g. (PERSON, RIDE, HORSE), with corresponding bounding boxes in the images~\cite{gupta2015visual}. We use the subset of V-COCO obtained by discarding the actions that have no argument, as described in Section \ref{sec:recCOCO}. This allows for head-to-head comparison with REC-COCO. 

\paragraph{REC-COCO} It contains the same examples above, but the subject, object and relation tokens have been extracted from the captions after the automatic link to the  concept triplets, e.g. (\emph{woman}, \emph{riding}, \emph{horse}). 

\paragraph{Visual Genome} This dataset \cite{Krishna2017} also contains manually annotated ($S,R,O$) concept triplets with corresponding bounding boxes in the images. The images and annotations are independent of those in V-COCO (and therefore REC-COCO). We use two variants of this dataset: the $378k$ version is the same used in \cite{collell2018acquiring}, where all instances containing explicit relations are discarded. We further reduce this dataset to have the same size as REC-COCO. The subset is created by randomly selecting triplets of the $21$ most used actions and the $67$ most used entities. %Those two variants are used to assess the first feature of the task, i.e. the difficulty.

\vspace{0.3cm}
Table \ref{dataset_table} shows the results of the system in each dataset. Regarding the effect of the automatic mapping, the table shows that using the caption tokens (i.e. REC-COCO) instead of ontology concept triplets (i.e. V-COCO) yields  better results in three of the evaluation metrics and worse in the other three metrics, so we can conclude that they are  comparable. These results show that the possible errors introduced when aligning triplets to caption tokens is relatively low, and that REC-COCO is overall a valid dataset for inferring spatial relations from triplets.

The results are better than those obtained using a subset of Visual Genome of comparable size, although a larger training dataset ($378k$) yields better results overall. Although the results of different datasets can not be directly compared, they can give insights regarding the task difficulty. In this regard, the table shows that the task proposed by REC-COCO is comparable in difficulty to the triplets presented in Visual Genome. As Visual Genome is a well established dataset, we think these results are relevant. 

\begin{table*}[t]
\begin{center}
\begin{tabular}{|l|cccccc|}
\hline
Input & $acc_y$         & $F1_y$          & $r_x$           & $r_y$           & $R^2$           & IoU \\
\hline
Conceptual relation       & 75.6          & 75.4          & 78.3          & 63.4          & 51.7          & 14.9          \\
Relation token          & \textbf{77.9} & \textbf{77.7} & 70.4         & \textbf{67.6}  & 47.3          & 12.1 \\
\hline
Caption & 77.6          & \textbf{77.7}          & \textbf{82.6} & 65.2          & \textbf{59.3} & \textbf{16.4} \\
\hline
\end{tabular}
{\caption{Evaluating the contribution of captions. Performance for different inputs: manual conceptual relations (V-COCO subset), the relation token in the caption (as linked when deriving REC-COCO) and full captions (REC-COCO). Results are fully comparable, as they only differ in the input used, and show the performance gains when using captions. See text for more details. }\label{table2}}
\end{center}
\end{table*}

\subsection{Evaluating the contribution of captions}

Table \ref{table2} shows the results which confirm the hypothesis in this paper. The top row shows the results when the system ignores the caption and uses instead the manually extracted conceptual relation in V-COCO. The second row shows the results of the same system when using the automatically mapped relation token in the input\footnote{These results are the same as the rows labeled with V-COCO and REC-COCO in Table \ref{dataset_table}.}. The bottom row shows the results for the system when using the full caption encoded using BERT, ignoring which is the relevant relation\footnote{The results for alternative caption encoders are shown below.}.  The best results in all metrics except $acc_y$ and $r_y$ are obtained when using the caption. Indeed, this model achieves the best $R^2$ and IoU, the metrics which are best for evaluating spatial relations, since they are continuous and consider both $x$ and $y$ axis. These results confirm our hypothesis: (1) it is possible to infer the size and location of an object with respect to a given subject directly from the caption; (2) the use of full text allows to place the object better than using a manually extracted relation. The improvement obtained by the use of full captions with respect to using the relation token alone reflects that the motivation was correct (cf. Figure \ref{problem}).

\subsection{Evaluating caption encoders}

Table \ref{table1} shows the results of our model on REC-COCO for different caption encoders (cf. Section \ref{sec:models-inferr-spat}). As expected, the simplest model (AVG) yields the worst results across all metrics, and the better results of BiLSTM show that this sentence encoder is able to profit from modelling word order in order to learn a more effective caption representation. In the bottom row, the use of transformers pre-trained in a masked language modelling task to encode the caption (BERT) yields the best results for all metrics except $acc_y$ and $F1_y$. The fact that these results agree with those obtained by the community on sentence encoding problems across multiple tasks \cite{devlin2018bert} serves as indirect confirmation that REC-COCO is a well-designed dataset, and that full captions contain information which is relevant for inducing spatial relations.

\begin{table*}[t]
\begin{center}
\begin{tabular}{|l|cccccc|}
\hline
 Encoder & $acc_y$         & $F1_y$          & $r_x$           & $r_y$           & $R^2$           & IoU
                                                                                                                                         \\
\hline
AVG     & 77.8          & 77.7          & 80.0          & 60.4          & 53.8          & 13.8          \\
BiLSTM  & \textbf{79.4}          & \textbf{79.5}          & 80.0          & 64.9          & 56.6          & 15.0          \\
BERT    & 77.6          & 77.7          & \textbf{82.6} & \textbf{65.2}          & \textbf{59.3} & \textbf{16.4} \\
\hline
\end{tabular}
{\caption{Evaluating caption encoders. Performance on REC-COCO for different caption encoders (cf. Section \ref{sec:models-inferr-spat})}\label{table1}}
\end{center}
\end{table*}

\begin{table*}[t]
\begin{center}
\begin{tabular}{|l|cccccc|}
\hline
Input & $acc_y$         & $F1_y$          & $r_x$           & $r_y$           & $R^2$           & IoU                               \\
\hline
Caption    & 77.6          & 77.7          & \textbf{82.6} & 65.2          & \textbf{59.3} & \textbf{16.4} \\
%\hline
Caption+relation     & \textbf{79.5} & \textbf{79.6} & \textbf{82.6} & \textbf{66.1}          & 58.6          & 14.9          \\

Caption-SO      & 65.0          & 65.1          & 77.4          & 20.1          & 18.8          & 2.3\\
\hline
\end{tabular}
{\caption{Analysis of combined inputs. Performance on REC-COCO for different inputs in each row: caption, caption plus relation, 
%relation token instead of caption, 
caption without subject and object information. }\label{table3}}
\end{center}
\end{table*}

\subsection{Analysis of combined inputs}

In order to understand the contribution of each possible input, we tried several additional combinations and ablations, as shown in Table \ref{table3}. In the first row we show the results when using the BERT encoding over captions, already reported in Table \ref{table2} and repeated here for easier comparison. In the second row we show the results when extending the input to consider the embedding of the relation in addition to the caption, with no clear improvement, as the results only improve slightly in the binary above/below metric ($acc_y$ and $F1_y$), with lower performance in other metrics. This result shows that the caption encoder is able to represent the relevant information regarding the relation between subject and object, without the need of an additional explicit signal.

The third row shows the results when our model (cf. Figure \ref{modelfig} in Section \ref{sec:models-inferr-spat}) does not receive any information about which are the subject and object. The clear decrease in performance shows that the model is not learning hidden biases in the data, and that the results of our model are sensible. From another perspective, it also shows that the captions in our dataset are complex and describe multiple relations between different subjects and objects. 

All in all, the results validate our hypothesis that the information conveyed in captions is complementary to the structured information, and that the unstructured information is particularly useful when important information is missing from the triplets.

\subsection{Error Analysis}
\begin{figure*}[t!]
    \centerline{\includegraphics[height=2.5in, width=5.8in]{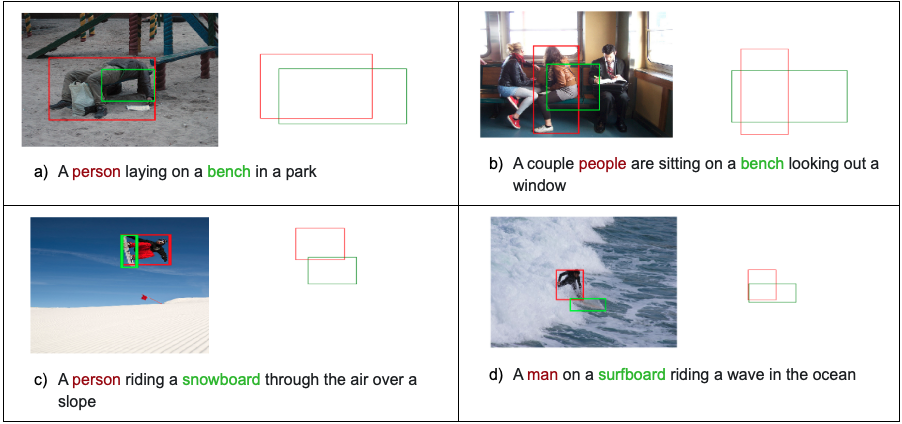}}
\caption{Four examples of spatial relations inferred by our method (green bounding box on white background, right side) that do not match the gold standard (green bounding box in the image, left side). Best viewed in color.} \label{error_adibideak}
\end{figure*}

The MS-COCO dataset contains complex scenes with many entities in diverse contexts, which makes spatial relations prediction very challenging. Even the context provided by captions may be insufficient to properly identify the spatial relations of some images. Figure \ref{error_adibideak} shows examples of system predictions that do not agree with the ground truth. The a) example shows a difficult scene where the caption does not provide enough information about the scene. Note that, although wrong, the system prediction corresponds more or less to prototypical spatial arrangements between the entities mentioned in the scene, which would probably agree with the spatial relations that a typical person would draw.

The b) example shows incorrectly tagged entities. For example, the bench in the image is larger than the tagged bounding box. But the model prediction for the bounding box suggests that it knows that a bench is usually larger than a person. Further, when we compare a) and b) examples we see that our model is also able to differentiate when a person is laying or is sitting on a bench. When it is laying, the bench is roughly equal in size to the person, but it is larger in the $x$ axis when the person is sitting on a bench. This is something interesting, because it shows the ability to learn common sense from the raw text and visual information, like humans do.

The c) example in Figure \ref{error_adibideak} shows another difficult scene, where the person is jumping with his/her snowboard. The position of the person is not the usual one (on top of the board). This is not fully described in the caption, and our model infers that the person is on the board. Once again, it would be interesting to see what humans would draw given the caption. The d) example is also complicated, since it shows the occlusion of the object, which our model cannot handle properly. In that case, the surfboard is well located (under the person), but its size is larger than in the image. However, it is worth to note that the bounding box predicted for the surfboard is not long enough (given the size of the person, the surfboard should be longer in the $x$ axis, if it was not occluded). It might be that the "riding a wave" expression made the model infer that a part of the board is actually occluded. %This is just a guess that should be further explored.

\section{Conclusions}

In this paper, we show that using the full textual descriptions of images improves the ability to model the spatial relationships between entities. Previous research has focused on using structured concept triplets which include an ontological representation of the relation, but we show that the caption contains additional useful information which our system uses effectively to improve results. Our experiments are based on REC-COCO, a new dataset that we have automatically derived from   MS-COCO and V-COCO containing associations between the words in the caption and bounding boxes in images. Although there is some lose of information when moving from the ontological concept triplets to the corresponding textual triplet as mentioned in the textual caption, the use of the full caption yields the best results. Furthermore, we see that the improvement also holds without explicitly specifying the relation token in the caption, which shows that our system is able to automatically place entities relative to others without any additional manual annotation. The system is thus able to figure out the relation and the relevant contextual information from the textual caption. Our error analysis shows that even in the case of examples where the system output gets low scores, the system often guesses prototypical locations and sizes, which we think reflect common sense knowledge about the scenes.

In order to place an object according to a caption, our system needs to take as reference the size and location of another object. In the future, we would like to explore techniques to infer, from a caption describing a scene, which entities need to be depicted and their respective location and sizes in the image. In addition, recent multimodal transformers like X-LXMERT \cite{cho2020xlxmert} could be used to improve the encoding of captions, taking advantage of the visual grounding previously learned by the model. Finally, given that our work shows that it is not necessary to manually annotate the relation between entities for satisfactory results, large collections like MS-COCO which include captions and bounding boxes can be readily used to train and test systems with the ability to decide which entities from a caption need to be depicted.  

\section*{Acknowledgements}
Aitzol Elu has been supported by a ETORKIZUNA ERAIKIZ grant from the Provincial Council of Gipuzkoa. This research has been partially funded by the Basque Government excellence research group (IT1343-19), the Spanish MINECO (FuturAAL RTI2018-101045-B-C21, DeepReading RTI2018-096846-B-C21 MCIU/AEI/FEDER, UE), Project BigKnowledge (Ayudas Fundación BBVA a equipos de investigación científica 2018), and the NVIDIA GPU grant program.

\section*{References}

\bibliographystyle{elsarticle-num}
\bibliography{bibliography}

\begin{thebibliography}{10}
\expandafter\ifx\csname url\endcsname\relax
  \def\url#1{\texttt{#1}}\fi
\expandafter\ifx\csname urlprefix\endcsname\relax\def\urlprefix{URL }\fi
\expandafter\ifx\csname href\endcsname\relax
  \def\href#1#2{#2} \def\path#1{#1}\fi

\bibitem{mogadala2019trends}
A.~Mogadala, M.~Kalimuthu, D.~Klakow, Trends in integration of vision and
  language research: A survey of tasks, datasets, and methods, arXiv preprint
  arXiv:1907.09358.

\bibitem{vandume2010}
B.~D.~V. Dume, Extracting implicit knowledge from text, Ph.D. thesis,
  University of Rochester (2010).

\bibitem{reed2016generative}
S.~Reed, Z.~Akata, X.~Yan, L.~Logeswaran, B.~Schiele, H.~Lee,
  \href{http://proceedings.mlr.press/v48/reed16.html}{Generative adversarial
  text to image synthesis}, in: M.~F. Balcan, K.~Q. Weinberger (Eds.),
  Proceedings of The 33rd International Conference on Machine Learning, Vol.~48
  of Proceedings of Machine Learning Research, PMLR, New York, New York, USA,
  2016, pp. 1060--1069.
\newline\urlprefix\url{http://proceedings.mlr.press/v48/reed16.html}

\bibitem{bagherinezhad2016elephants}
H.~Bagherinezhad, H.~Hajishirzi, Y.~Choi, A.~Farhadi, Are elephants bigger than
  butterflies? reasoning about sizes of objects, in: Thirtieth AAAI Conference
  on Artificial Intelligence, 2016.

\bibitem{collell2018acquiring}
G.~Collell, L.~Van~Gool, M.-F. Moens, Acquiring common sense spatial knowledge
  through implicit spatial templates, in: Thirty-Second AAAI Conference on
  Artificial Intelligence, 2018.

\bibitem{gupta2015visual}
S.~Gupta, J.~Malik, Visual semantic role labeling, arXiv preprint
  arXiv:1505.04474.

\bibitem{huang2019sketchforme}
F.~Huang, J.~F. Canny,
  \href{http://doi.acm.org/10.1145/3332165.3347878}{Sketchforme: Composing
  sketched scenes from text descriptions for interactive applications}, in:
  Proceedings of the 32Nd Annual ACM Symposium on User Interface Software and
  Technology, UIST '19, ACM, New York, NY, USA, 2019, pp. 209--220.
\newblock \href {http://dx.doi.org/10.1145/3332165.3347878}
  {\path{doi:10.1145/3332165.3347878}}.
\newline\urlprefix\url{http://doi.acm.org/10.1145/3332165.3347878}

\bibitem{Krishna2017}
R.~Krishna, Y.~Zhu, O.~Groth, J.~Johnson, K.~Hata, J.~Kravitz, S.~Chen,
  Y.~Kalantidis, L.-J. Li, D.~A. Shamma, M.~S. Bernstein, L.~Fei-Fei, Visual
  genome: Connecting language and vision using crowdsourced dense image
  annotations, Int. J. Comput. Vision 123~(1) (2017) 32--73.

\bibitem{malinowski2014pooling}
M.~Malinowski, M.~Fritz, A pooling approach to modelling spatial relations for
  image retrieval and annotation, arXiv preprint arXiv:1411.5190.

\bibitem{lin2014microsoft}
T.-Y. Lin, M.~Maire, S.~Belongie, J.~Hays, P.~Perona, D.~Ramanan,
  P.~Doll{\'a}r, C.~L. Zitnick, Microsoft coco: Common objects in context, in:
  European conference on computer vision, Springer, 2014, pp. 740--755.

\bibitem{guadarrama2013grounding}
S.~Guadarrama, L.~Riano, D.~Golland, D.~Go, Y.~Jia, D.~Klein, P.~Abbeel,
  T.~Darrell, et~al., Grounding spatial relations for human-robot interaction,
  in: 2013 IEEE/RSJ International Conference on Intelligent Robots and Systems,
  IEEE, 2013, pp. 1640--1647.

\bibitem{hinz2019generating}
T.~Hinz, S.~Heinrich, S.~Wermter,
  \href{https://openreview.net/forum?id=H1edIiA9KQ}{Generating multiple objects
  at spatially distinct locations}, in: International Conference on Learning
  Representations, 2019.
\newline\urlprefix\url{https://openreview.net/forum?id=H1edIiA9KQ}

\bibitem{jyothi2019layoutvae}
A.~A. Jyothi, T.~Durand, J.~He, L.~Sigal, G.~Mori, Layoutvae: Stochastic scene
  layout generation from a label set, in: The IEEE International Conference on
  Computer Vision (ICCV), 2019.

\bibitem{Kruijff2007}
G.-J. Kruijff, H.~Zender, P.~Jensfelt, H.~Christensen, Situated dialogue and
  spatial organization: What, where and why?, International Journal of Advanced
  Robotic Systems 4.
\newblock \href {http://dx.doi.org/10.5772/5701} {\path{doi:10.5772/5701}}.

\bibitem{platonov2018computational}
G.~Platonov, L.~Schubert, Computational models for spatial prepositions, in:
  Proceedings of the First International Workshop on Spatial Language
  Understanding, 2018, pp. 21--30.

\bibitem{goodfellow2014generative}
I.~Goodfellow, J.~Pouget-Abadie, M.~Mirza, B.~Xu, D.~Warde-Farley, S.~Ozair,
  A.~Courville, Y.~Bengio, Generative adversarial nets, in: Advances in neural
  information processing systems, 2014, pp. 2672--2680.

\bibitem{Xu_2018_CVPR}
T.~Xu, P.~Zhang, Q.~Huang, H.~Zhang, Z.~Gan, X.~Huang, X.~He, Attngan:
  Fine-grained text to image generation with attentional generative adversarial
  networks, in: The IEEE Conference on Computer Vision and Pattern Recognition
  (CVPR), 2018.

\bibitem{NIPS2016_6111}
S.~E. Reed, Z.~Akata, S.~Mohan, S.~Tenka, B.~Schiele, H.~Lee,
  \href{http://papers.nips.cc/paper/6111-learning-what-and-where-to-draw.pdf}{Learning
  what and where to draw}, in: D.~D. Lee, M.~Sugiyama, U.~V. Luxburg, I.~Guyon,
  R.~Garnett (Eds.), Advances in Neural Information Processing Systems 29,
  Curran Associates, Inc., 2016, pp. 217--225.
\newline\urlprefix\url{http://papers.nips.cc/paper/6111-learning-what-and-where-to-draw.pdf}

\bibitem{48087}
S.~{Hong}, D.~{Yang}, J.~{Choi}, H.~{Lee}, Inferring semantic layout for
  hierarchical text-to-image synthesis, in: 2018 IEEE/CVF Conference on
  Computer Vision and Pattern Recognition, 2018, pp. 7986--7994.
\newblock \href {http://dx.doi.org/10.1109/CVPR.2018.00833}
  {\path{doi:10.1109/CVPR.2018.00833}}.

\bibitem{li2019objectdriven}
W.~Li, P.~Zhang, L.~Zhang, Q.~Huang, X.~He, S.~Lyu, J.~Gao, Object-driven
  text-to-image synthesis via adversarial training, in: The IEEE Conference on
  Computer Vision and Pattern Recognition (CVPR), 2019.

\bibitem{forbes-choi-2017-verb}
M.~Forbes, Y.~Choi, \href{https://www.aclweb.org/anthology/P17-1025}{Verb
  physics: Relative physical knowledge of actions and objects}, in: Proceedings
  of the 55th Annual Meeting of the Association for Computational Linguistics
  (Volume 1: Long Papers), Association for Computational Linguistics,
  Vancouver, Canada, 2017, pp. 266--276.
\newblock \href {http://dx.doi.org/10.18653/v1/P17-1025}
  {\path{doi:10.18653/v1/P17-1025}}.
\newline\urlprefix\url{https://www.aclweb.org/anthology/P17-1025}

\bibitem{yang-etal-2018-extracting}
Y.~Yang, L.~Birnbaum, J.-P. Wang, D.~Downey,
  \href{https://www.aclweb.org/anthology/P18-2102}{Extracting commonsense
  properties from embeddings with limited human guidance}, in: Proceedings of
  the 56th Annual Meeting of the Association for Computational Linguistics
  (Volume 2: Short Papers), Association for Computational Linguistics,
  Melbourne, Australia, 2018, pp. 644--649.
\newblock \href {http://dx.doi.org/10.18653/v1/P18-2102}
  {\path{doi:10.18653/v1/P18-2102}}.
\newline\urlprefix\url{https://www.aclweb.org/anthology/P18-2102}

\bibitem{silberer_uijlings_lapata_2018}
C.~Silberer, J.~Uijlings, M.~Lapata, Understanding visual scenes, Natural
  Language Engineering 24~(3) (2018) 441–465.
\newblock \href {http://dx.doi.org/10.1017/S1351324918000104}
  {\path{doi:10.1017/S1351324918000104}}.

\bibitem{wang2020learning}
J.~Wang, W.~Wang, L.~Wang, Z.~Wang, D.~D. Feng, T.~Tan, Learning visual
  relationship and context-aware attention for image captioning, Pattern
  Recognition 98 (2020) 107075.

\bibitem{aditya2019spatial}
S.~Aditya, R.~Saha, Y.~Yang, C.~Baral, Spatial knowledge distillation to aid
  visual reasoning, in: 2019 IEEE Winter Conference on Applications of Computer
  Vision (WACV), IEEE, 2019, pp. 227--235.

\bibitem{bai2020decomvqanet}
Z.~Bai, Y.~Li, M.~Wo{\'z}niak, M.~Zhou, D.~Li, Decomvqanet: Decomposing visual
  question answering deep network via tensor decomposition and regression,
  Pattern Recognition 110 (2020) 107538.

\bibitem{davidov2010extraction}
D.~Davidov, A.~Rappoport, Extraction and approximation of numerical attributes
  from the web, in: Proceedings of the 48th Annual Meeting of the Association
  for Computational Linguistics, Association for Computational Linguistics,
  2010, pp. 1308--1317.

\bibitem{narisawa2013204}
K.~Narisawa, Y.~Watanabe, J.~Mizuno, N.~Okazaki, K.~Inui, Is a 204 cm man tall
  or small? acquisition of numerical common sense from the web, in: Proceedings
  of the 51st Annual Meeting of the Association for Computational Linguistics
  (Volume 1: Long Papers), 2013, pp. 382--391.

\bibitem{tandon2014acquiring}
N.~Tandon, G.~De~Melo, G.~Weikum, Acquiring comparative commonsense knowledge
  from the web, in: Twenty-Eighth AAAI Conference on Artificial Intelligence,
  2014.

\bibitem{DBLP:journals/corr/abs-1906-01327}
Y.~Elazar, A.~Mahabal, D.~Ramachandran, T.~Bedrax-Weiss, D.~Roth, How large are
  lions? inducing distributions over quantitative attributes, in: Proceedings
  of the 57th Annual Meeting of the Association for Computational Linguistics,
  Association for Computational Linguistics, 2019.
\newblock \href {http://dx.doi.org/10.18653/v1/p19-1388}
  {\path{doi:10.18653/v1/p19-1388}}.

\bibitem{devlin2018bert}
J.~Devlin, M.-W. Chang, K.~Lee, K.~Toutanova,
  \href{https://www.aclweb.org/anthology/N19-1423}{{BERT}: Pre-training of deep
  bidirectional transformers for language understanding}, in: Proceedings of
  the 2019 Conference of the North {A}merican Chapter of the Association for
  Computational Linguistics: Human Language Technologies, Volume 1 (Long and
  Short Papers), Association for Computational Linguistics, Minneapolis,
  Minnesota, 2019, pp. 4171--4186.
\newblock \href {http://dx.doi.org/10.18653/v1/N19-1423}
  {\path{doi:10.18653/v1/N19-1423}}.
\newline\urlprefix\url{https://www.aclweb.org/anthology/N19-1423}

\bibitem{lu2019vilbert}
J.~Lu, D.~Batra, D.~Parikh, S.~Lee, Vilbert: Pretraining task-agnostic
  visiolinguistic representations for vision-and-language tasks, in: Advances
  in Neural Information Processing Systems, 2019, pp. 13--23.

\bibitem{li2019visualbert}
L.~H. Li, M.~Yatskar, D.~Yin, C.-J. Hsieh, K.-W. Chang, Visualbert: A simple
  and performant baseline for vision and language, arXiv preprint
  arXiv:1908.03557.

\bibitem{lin2020interbert}
J.~Lin, A.~Yang, Y.~Zhang, J.~Liu, J.~Zhou, H.~Yang, Interbert:
  Vision-and-language interaction for multi-modal pretraining, arXiv preprint
  arXiv:2003.13198.

\bibitem{cho2020xlxmert}
J.~Cho, J.~Lu, D.~Schwenk, H.~Hajishirzi, A.~Kembhavi, {X-LXMERT}: {P}aint,
  {C}aption and {A}nswer {Q}uestions with {M}ulti-{M}odal {T}ransformers, in:
  Proceedings of the 2020 Conference on Empirical Methods in Natural Language
  Processing (EMNLP), Association for Computational Linguistics, Online, 2020,
  pp. 8785--8805.
\newblock \href {http://dx.doi.org/10.18653/v1/2020.emnlp-main.707}
  {\path{doi:10.18653/v1/2020.emnlp-main.707}}.

\bibitem{graves2013speech}
A.~Graves, A.-r. Mohamed, G.~Hinton, Speech recognition with deep recurrent
  neural networks, in: 2013 IEEE international conference on acoustics, speech
  and signal processing, IEEE, 2013, pp. 6645--6649.

\bibitem{draper1998applied}
N.~R. Draper, H.~Smith, Applied regression analysis, Vol. 326, John Wiley \&
  Sons, 1998.

\bibitem{everingham2015pascal}
M.~Everingham, S.~A. Eslami, L.~Van~Gool, C.~K. Williams, J.~Winn,
  A.~Zisserman, The pascal visual object classes challenge: A retrospective,
  International journal of computer vision 111~(1) (2015) 98--136.

\bibitem{Pennington14glove:global}
J.~Pennington, R.~Socher, C.~Manning,
  \href{https://www.aclweb.org/anthology/D14-1162}{{G}lo{V}e: Global vectors
  for word representation}, in: Proceedings of the 2014 Conference on Empirical
  Methods in Natural Language Processing ({EMNLP}), Association for
  Computational Linguistics, Doha, Qatar, 2014, pp. 1532--1543.
\newblock \href {http://dx.doi.org/10.3115/v1/D14-1162}
  {\path{doi:10.3115/v1/D14-1162}}.
\newline\urlprefix\url{https://www.aclweb.org/anthology/D14-1162}

\end{thebibliography}

\end{document}